\documentclass[runningheads]{llncs}
\usepackage{graphicx}

\usepackage{tikz}
\usepackage{comment}
\usepackage{amsmath,amssymb} 
\usepackage{color}

\usepackage[accsupp]{axessibility}  

\usepackage{booktabs}
\usepackage{enumitem}

\usepackage{lipsum}
\usepackage{acronym}
\usepackage{algpseudocode}
\usepackage{algorithm}
\usepackage{pifont}
\newcommand{\cmark}{\ding{51}}%
\usepackage{amsfonts}
\usepackage{siunitx}
\usepackage{subfigure}
\usepackage{multirow}
\usepackage{boxhandler}
\usepackage{tabularx}
\usepackage{adjustbox}
\usepackage{epstopdf}
\usepackage{colortbl}
\definecolor{mygray}{gray}{.9}

\usepackage{wrapfig}
\usepackage{capt-of}
\usepackage[rightcaption]{sidecap} 

\usepackage[pagebackref,breaklinks,colorlinks]{hyperref}

\usepackage[capitalize]{cleveref}
\crefname{section}{Sec.}{Secs.}
\Crefname{section}{Section}{Sections}
\Crefname{table}{Table}{Tables}
\crefname{table}{Tab.}{Tabs.}

\newcommand{\ie}{\textit{i}.\textit{e}.}
\newcommand{\eg}{\textit{e}.\textit{g}.}

\begin{document}
\pagestyle{headings}
\mainmatter
\def\ECCVSubNumber{2280}  

\title{How to Synthesize a Large-Scale and Trainable Micro-Expression Dataset?}



\titlerunning{How to Synthesize a Large-Scale and Trainable Micro-Expression Dataset?}
%
\author{Yuchi Liu \inst{1} \and
Zhongdao Wang\inst{2} \and
Tom Gedeon\inst{1} \and
Liang Zheng \inst{1}}
\authorrunning{Y. Liu, et al.}
%
\institute{Australian National University, Canberra, Australia, \\
\email{\{firstname.lastname\}@anu.edu.au} \and
Tsinghua University, Beijing, China \email{wcd17@mails.tsinghua.edu.cn}\\
\url{https://github.com/liuyvchi/MiE-X}
}
\maketitle

\begin{abstract}
This paper does not contain technical novelty but introduces our key discoveries in a data generation protocol, a database and insights. We aim to address the lack of large-scale datasets in micro-expression (MiE) recognition due to the prohibitive cost of data collection, which renders large-scale training less feasible. To this end, we develop a protocol to automatically synthesize large scale MiE training data that allow us to train improved recognition models for real-world test data. Specifically, we discover three types of Action Units (AUs) that can constitute trainable MiEs. These AUs come from real-world MiEs, early frames of macro-expression videos, and the relationship between AUs and expression categories defined by human expert knowledge. With these AUs, our protocol then employs large numbers of face images of various identities and an off-the-shelf face generator for MiE synthesis, yielding the MiE-X dataset. MiE recognition models are trained or pre-trained on MiE-X and evaluated on real-world test sets, where very competitive accuracy is obtained. Experimental results not only validate the effectiveness of the discovered AUs and MiE-X dataset but also reveal some interesting properties of MiEs: they generalize across faces, are close to early-stage macro-expressions, and can be manually defined\footnote{This work was supported by the ARC Discovery Early Career Researcher Award (DE200101283) and the ARC Discovery Project (DP210102801).}.

\end{abstract}

\section{Introduction}

Micro-Expressions (MiEs) are transient facial expressions that typically last for $0.04$ to $0.2$  seconds~\cite{matsumoto2008culture,ekman1997face}. Unlike conventional facial expressions (or Macro-Expressions, MaEs) that last for longer than $0.2$ seconds, MiEs are involuntary. They are difficult to be pretended, and thus more capable of revealing people's genuine emotions. MiE recognition underpins various valuable applications such as lie detection, criminal justice, and psychological consultation.

\begin{figure}[t]
 \begin{minipage}[c]{0.4\textwidth}
    \includegraphics[width=\textwidth]{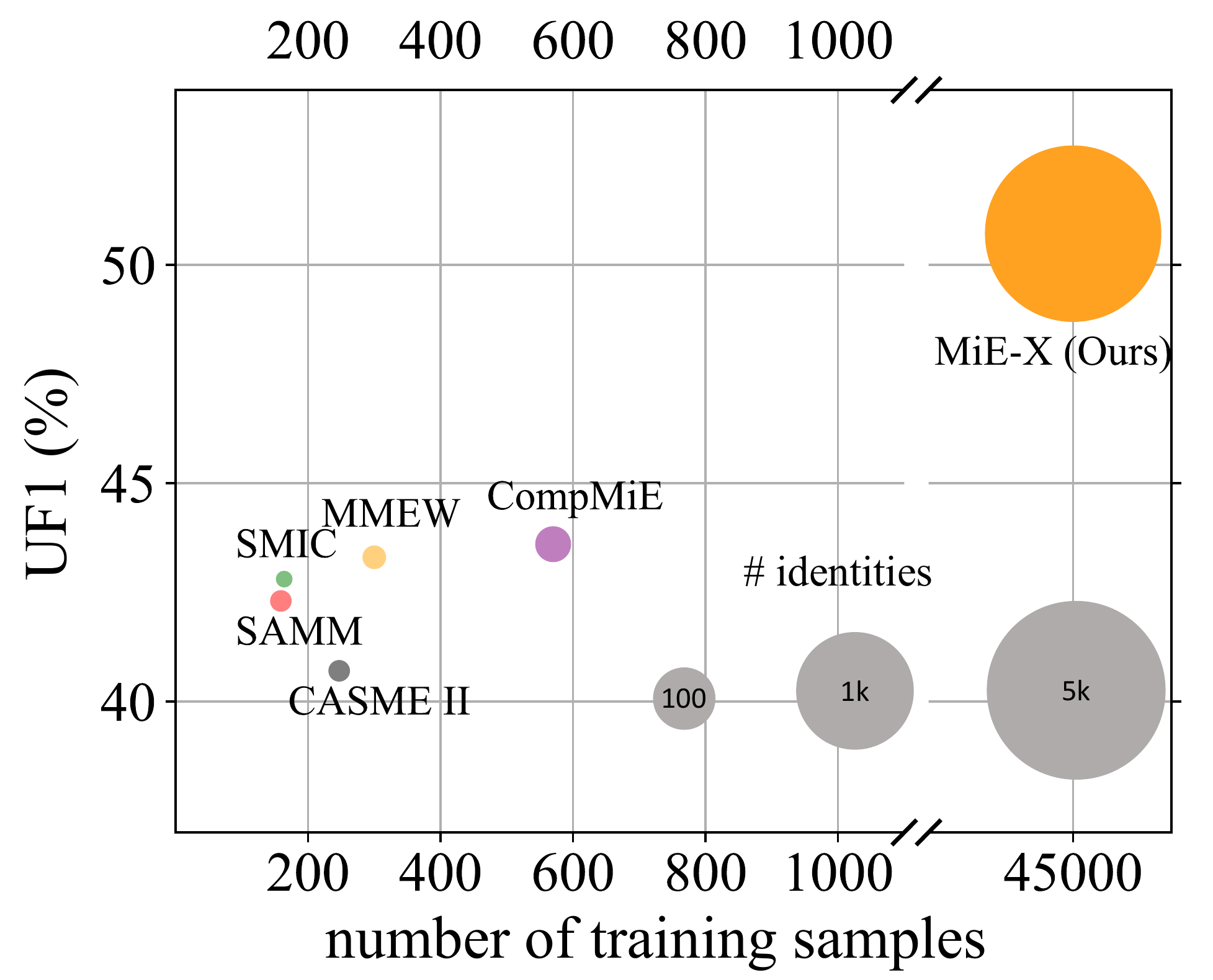}
  \end{minipage}
  \hfill
\begin{minipage}[c]{0.56\textwidth}
    \caption{We present a large-scale synthetic MiE training dataset, {MiE-X}, created by the proposed protocol. It is two magnitudes larger than existing real MiE recognition datasets in terms of number of MiE samples and number of identities. Compared with existing real-world MiE datasets, MiE-X allows the MiE classifier~\cite{Liu2018A} to achieve consistently higher accuracy evaluated on the real-world MiE dataset CompMiE~\cite{see2019megc}.}
  \label{fig:compare_datasets}
  \end{minipage}
\end{figure}

The difficulty in collecting and labeling MiEs poses huge challenges in building MiE recognition datasets~\cite{ben2021video}. First, collecting \emph{involuntary} MiEs is strenuous, even in a controlled environment~\cite{ben2021video}. Unlike MaEs, which participants can easily ``perform'', MiEs are too vague and subtle to precisely interpret. Second, correctly labeling MiEs is difficult. It usually requires domain knowledge from psychology experts, and oftentimes even experts cannot guarantee a high accuracy of annotations. 
As a consequence, scales of existing MiE recognition datasets are severely limited: they typically consist of a few hundreds of samples from dozens of identities (refer Fig.~\ref{fig:compare_datasets} for an illustrative summary). Shortage of training data would compromise the development of MiE recognition algorithms.

In this work, we aim to address the data shortage issue by proposing a useful protocol for \emph{synthesizing} MiEs. This protocol has three steps. First, we conveniently obtain a large number of faces from existing face datasets.  
Second, we compute sensible AUs.
Third, we employ a conditional generative model to ``add'' MiEs onto these faces. Conditional facial expression generation is a well-studied problem, and we adopt an off-the-shelf algorithm, GANimation~\cite{pumarola2018ganimation}, which employs coefficients of Action Units (AUs) as the generative conditions. 

 
At the core of this synthesis protocol, we contribute in finding three types of AUs helpful in the second step. The \textbf{first} type, intuitively, are AUs extracted from real-world, annotated MiE datasets. Specifically, we extract AU coefficients of annotated MiE samples and use these AU coefficients as conditions to transfer corresponding MiEs to faces of other identities.
The \textbf{second} type are AUs extracted from {early-stage MaEs}. The formation of macro-expressions consists of a process of facial muscle movements, and we find early stages of these movements usually share similar values of AUs to those of MiEs.
The \textbf{third} type are AU combinations given by expert knowledge. For example, human observations suggest that AU12 (\texttt{{Lip Corner Puller}}) is often activated when the subject is ``happy'', so we set AU12 to be slightly greater than $0$ when synthesizing a ``happy'' MiE. In this regard, this work is an early attempt to explore the underlying \emph{computational} mechanism of micro-expressions, and it would be of value for the community facilitating the understanding of micro-expressions and the design of learning algorithms.

Using the proposed three types of AUs, our protocol allows us to create a large-scale synthetic dataset, \textbf{MiE-X}, to improve the accuracy of data-driven MiE recognition algorithms. As shown in Fig.~\ref{fig:compare_datasets}, MiE-X is two orders of magnitude larger than existing real-world datasets. Notably, despite being synthetic, MiE-X can be effectively used to train MiE recognition models. When the target application has the same label space as MiE-X, we can directly use MiE-X to train a recognition model, achieving competitive results to those trained on real-world data. Otherwise, MiE-X can be used for pre-training, and its pre-training quality outperforms ImageNet \cite{deng2009imagenet}. 
Our experiment shows that MiE-X consistently improves the accuracy of frame-based MiE recognition methods and a state-of-the-art video-based method. 



\begin{itemize}
   \item We introduce a large-scale MiE training dataset created by a useful protocol, for training MiE recognition models. The database will be released.
   \item We identify three types of AUs that allow for synthesizing trainable MiEs in the protocol. They are: 
   AUs extracted from real MiEs, mined from early-stage of MaEs and provided by human experts of facial expressions.
  \item Our experiments reveal interesting properties of MiEs: they generalize across identities, are close to early-stage MaEs, and can be manually defined.
\end{itemize}

\section{Related Work}\label{sec:relatedwork}
\textbf{Facial micro-expression recognition.}~Many MiE recognition systems use 
handcrafted features, such as 3DHOG \cite{polikovsky2009facial}, FDM~\cite{xu2017microexpression} and LBP-TOP \cite{zhao2007dynamic} descriptors. They describe facial texture patterns. 
Variants and extensions of LBP-TOP have also been proposed \cite{wang2014lbp,huang2015facial,huang2016spontaneous}. 
Afterwards, deep learning based solutions were proposed \cite{patel2016selective,kim2016micro,hao2017deep,peng2017dual,khor2018enriched,liong2018off}. Petal \emph{et al.}~\cite{patel2016selective} use the VGG model pretrained on ImageNet \cite{deng2009imagenet} and perform fine-tuning for MiE recognition. 
In ELRCN \cite{khor2018enriched}, the network input is enriched by the concatenation of the RGB image, optical flow and derivatives of optical flow~\cite{shreve2011macro}. To reduce computation cost and prevent overfitting, it is common to use representative frames as model input. For example, Peng \emph{et al.}~\cite{peng2018macro} and Li \emph{et al.}~\cite{li2018can} select the onset frame, apex frame and offset frame in each micro-expression video. Branches~\cite{Liu2018A} uses the onset and apex frame as model input. Following this practice, we focus on synthesising representative frames for MiEs. 

\textbf{Deep learning from synthetic data.}
Deep learning using synthetic data has drawn recent attention. Many works use graphic engines to generate virtual data and corresponding ground truths. Richter \emph{et al.} \cite{richter2016playing} use a 3D game engine to simulate training images with pixel-level label maps for semantic segmentation. In~\cite{sakaridis2018semantic}, prior human knowledge is used to constrain the distribution of synthetic target data. Tremblay \emph{et al.}~\cite{tremblay2018training} randomize the parameters of the simulator to force the model to handle large variations in object detection. Learning-based approaches~\cite{kar2019meta,ruiz2018learning,yao2020simulating} try to find the best parameter ranges in simulators so that the domain gap between generated content and the real-world data is minimized. Another line of works uses generative adversarial networks (GANs) to generate images for learning. For example, 
the label smoothing regularization technique is adopted for generated images \cite{zheng2017unlabeled}. Camstyle~\cite{zhong2018camera} trains camera-to-camera person appearance translation to generate new training data. 
CYCADA~\cite{hoffman2017cycada} reconstructs images and introduces semantic segmentation loss on these generated images to maintain consistent semantics. 

\textbf{Action Units (AUs) in facial analysis.}
Action Units are defined according to the Facial Action Coding System (FACS)~\cite{eckman1978facial}, which categorizes the fundamental facial muscles movements by their appearance on the face. Correlations between Action Units and emotions are widely discussed in literature~\cite{du2014compound,ekman1997face,polikovsky2013facial}. This work uses such correlations where we look for and validate effective AUs as generative conditions to synthesis realistic and trainable MiEs.

\section{Preliminaries}\label{sec:preliminary}
MiE recognition aims to classify emotion categories of a given face video clip. In practice, the video clips should be first processed by a \emph{spotting} algorithm to determine the onset (starting time), apex (time of the highest expression intensity) and offset (ending time) frames. In this work, we assume all data have been processed by spotting algorithms \cite{ben2021video,see2019megc} and focus on the recognition task.

Emotion labels in existing datasets are usually different, ranging from 3 to 8 categories. In this work, 
we use a unified and balanced label space to synthesize MiE-X. Specifically, during synthesis, we choose the most basic categories (\texttt{positive}, \texttt{negative}, \texttt{surprise}, as defined in MEGC) and merge other emotion labels into these three categories. If the label space in the target dataset is different from MiE-X, we need to fine-tune the model further. 

In the following sections, when mentioning {action units (AUs)}, we by default refer to the AU coefficient vector $\mathbf{z} \in [0,1]^d$. Each dimension in vector $\mathbf{z}$ indicates the intensity of a specific action unit. There are usually $d=17$ dimensions \cite{pumarola2018ganimation,baltrusaitis2018openface}. 

\section{Synthesizing Micro-Expressions}\label{sec:Synthesis}

\begin{figure}[t]
 \begin{minipage}[c]{0.5\textwidth}
    \includegraphics[width=\textwidth]{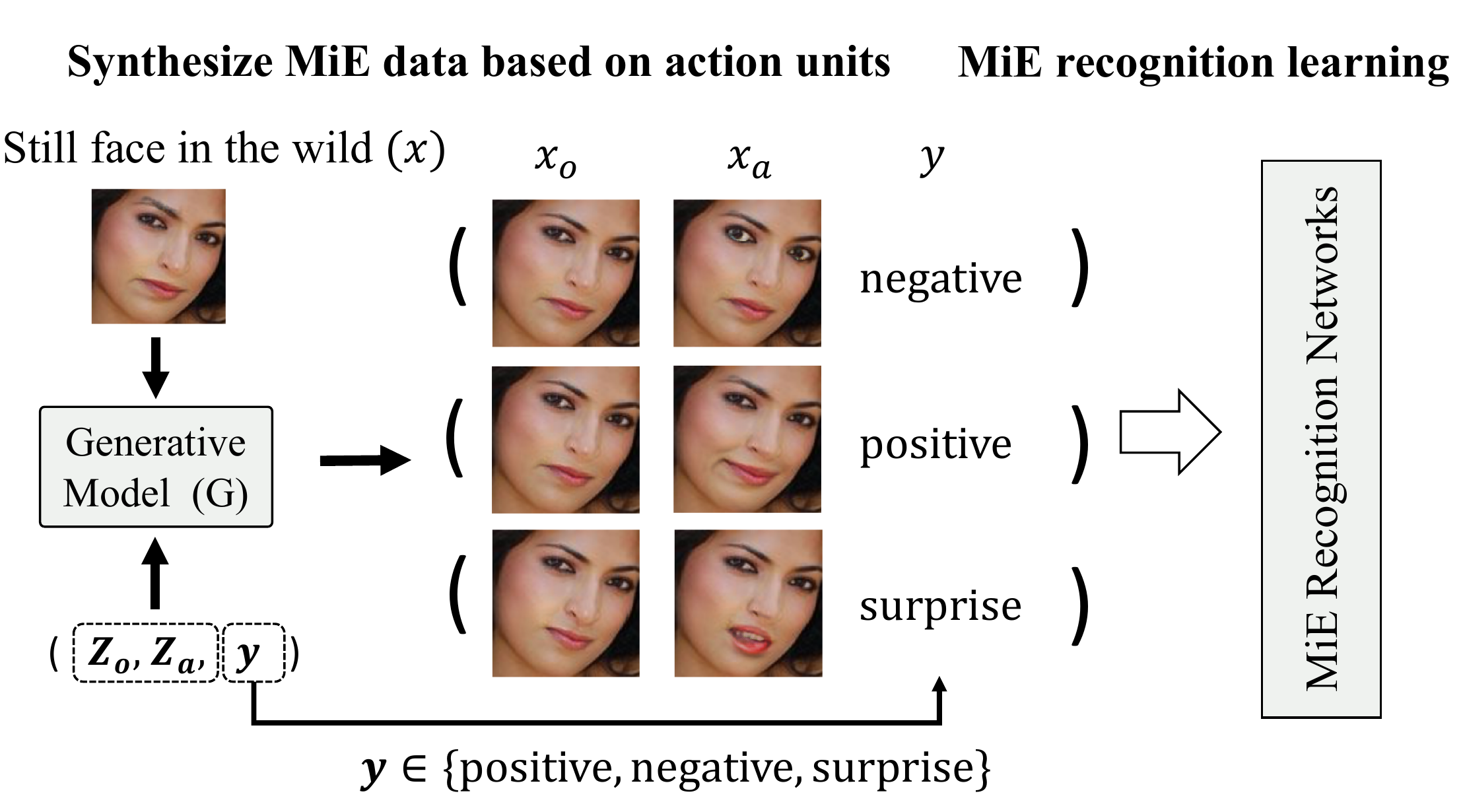}
  \end{minipage}
  \hfill
\begin{minipage}[c]{0.49\textwidth}
    \caption{
		Overview of the proposed protocol for synthesizing our MiE recognition dataset. We generate MiE samples (a triplet containing onset frame $\mathbf{x_o}$, apex frame $\mathbf{x_a}$ and the emotion label $y$) with a pretrained GANimation~\cite{pumarola2018ganimation} model \emph{G}, faces in the wild and AU vectors ($\mathbf{z}_o$, $\mathbf{z}_a$) introduced in Section~\ref{sec:AUs}.
    } \label{figure:overview}
  \end{minipage}
\end{figure}

\subsection{The Proposed Protocol}
Given a face image, an emotion label $y\in\{\texttt{positive}, \texttt{negative}, \texttt{surprise}\}$, and an onset-apex AU pair $(\mathbf{z_o}, \mathbf{z_a})$, our protocol uses GANimation \cite{pumarola2018ganimation} to generate an MiE sample consisting of two representative frames (refer Fig.~\ref{figure:overview}).

First, we randomly select an ``in-the-wild'' face image $\mathbf{x}$ from a large pool of identities (we use the EmotionNet~\cite{fabian2016emotionet} dataset) as the template face upon which we add MiEs. Then, we find an onset AU $\mathbf{z_o}$, an apex AU $\mathbf{z_a}$, and the corresponding emotion label $y$. A triplet of $(\mathbf{z_o}, \mathbf{z_a}, y)$ could be computed from three different sources, which are elaborated in Section~\ref{sec:AUs}. Finally, a conditional generative model $G$ is employed to transfer the onset and apex AUs to the template face $\mathbf{x}$, producing an onset frame $\mathbf{x_o} = G(\mathbf{x}, \mathbf{z_o})$ and an apex frame $\mathbf{x_a} = G(\mathbf{x}, \mathbf{z_a})$, whose emotion label is $y$ (same as the label of $\mathbf{x}$). Here, we adopt GANimation~\cite{pumarola2018ganimation} as $G$, which 
identity-preserving and only changes facial muscle movements. 
Training details of GANimation are provided in supp. materials. 

Please note that the protocol uses existing techniques and that we do not claim it as our main finding. 
Also note that we do not synthesize entire video sequences of MiEs, but only the onset (the beginning) and apex (most intensive) frames. The motivation is three-fold. First, a full MiE clip may contain up to 50 frames, so a dataset of full MiEs can be 25 times as large as a dataset of representative frames (2 frames per MiE). 
Second, recent literature on MiE recognition (\eg, \cite{peng2018macro,li2018can,Liu2018A}) indicate that using representative frames suffice to obtain very competitive accuracy. Last, synthesizing video sequences in a realistic way is much more challenging than static frames, requiring smooth motions and consistency over time. We leave video-level MiE generation to future work.

\subsection{Major Finding: Action Units That Constitute Trainable MiEs}\label{sec:AUs}  

In the protocol, we make the major contribution in finding three sources of AUs that are most helpful to define the onset and apex AUs, to be described below. 

\textbf{AUs extracted from real MiEs.} 
An intuitive source of MiE AUs are, of course, real-world MiE data. 
Assume we have a real-world MiE dataset with $M$ MiE videos, where each video is annotated with the onset and apex frames. 
For each video, we extract the onset and apex AUs and record the emotion label, forming a set of AUs $\mathcal{Z}^{\text{MiE}} = \{(\mathbf{z_o}^{(m)}, \mathbf{z_a}^{(m)}\}_{m=1}^{M}$ and labels $\mathcal{Y}^{\text{MiE}} = \{y^{(m)}\}_{m=1}^{M}$. Here, AU coefficients are extracted with the OpenFace toolkit~\cite{baltrusaitis2018openface}. 
When synthesizing MiEs with a certain emotion category based on $\mathbf{z}^{\text{MiE}}$, we randomly draw a pair of AUs from $\mathcal{Z}^{\text{MiE}}$ that have the desired emotion label. 

\emph{Discussion.} Despite being a valuable source of MiEs AUs, existing real-world MiE data are severely limited in size, so $\mathcal{Z}^{\text{MiE}}$ is far from being sufficient. If we had more MiE data, it would be interesting to further study whether our method can synthesize a better dataset. At this point, to include more MiE samples in our synthetic training set, we find another two AU sources below.

\textbf{AUs extracted from early-stage of real MaEs.}
Abundant MaE videos exist in the community, which have a similar set of emotion labels with MiE datasets. These MaE videos usually start from a neutral expression, leak subtle muscle movements in early frames, and present obvious expressions later. In our preliminary experiments, we observe that AUs extracted from early frames of MaE videos have similar values as those of MiE clips. 
This suggests that MiEs and \emph{early-stage} of MaEs have similar intensities in muscle movements, rendering the latter a potential source to simulate MiEs. 

In leveraging MaE videos as an AU source, we regard the first frame of MaE clips, which usually has a neutral expression, as our onset frame. The selection of the apex frame is more challenging. However, we empirically observe that existing MaE clips usually present MiE-liked AU intensities in the first half of the video. Therefore, we use two hyperparameters to find the apex frame approximately. Suppose an MaE clip has $n$ frames. An apex frame is randomly drawn from frame index $\lfloor\alpha \times n\rfloor$, where $\lfloor \cdot\rfloor$ rounds a number down to the nearest integer. The selections of $\alpha$ and $\beta$ are briefly discussed in Section~\ref{sec:exp:analysis}. 

\emph{Discussion.} 
Different MaE datasets may be different in the frame index of the onset and apex frames, so in practice we need to do a rapid scanning to roughly know them. But this process is usually quick, and importantly reliable, because 1) a certain dataset usually follows a stable pattern in terms of the onset and apex positions and 2) onset and apex states usually last for a while. As such, while this procedure requires a bit manual work, it is still very valuable considering the gain it brings (large-scale MiE data).

\begin{figure*}[t]  
    \begin{center}
        \includegraphics[width=\textwidth]{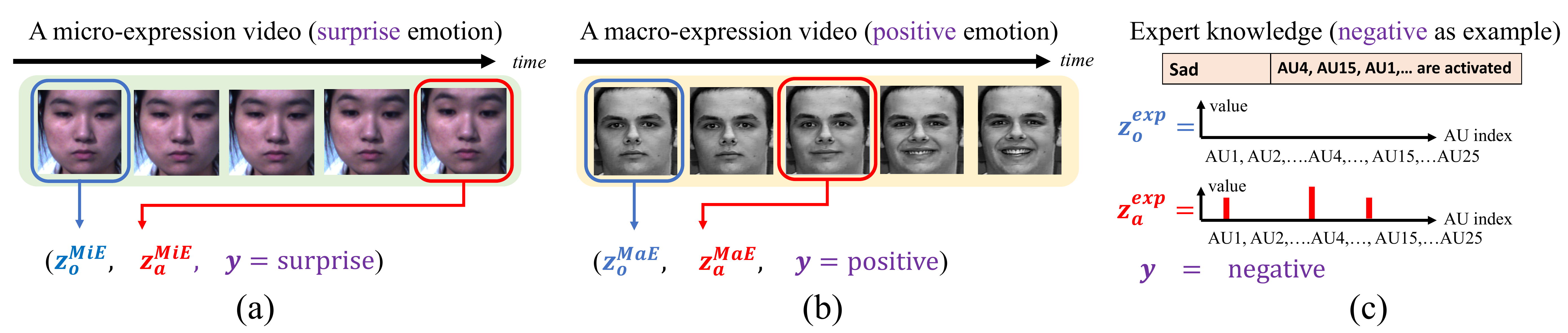}
        \caption{Examples of how to compute $\mathbf{z}^{\text{MiE}}$, $\mathbf{z}^{\text{MaE}}$ and $\mathbf{z}^{\text{exp}}$. \textbf{(a)} We compute $\mathbf{z}^{\text{MiE}}$ from representative frames (\ie, the onset frame and the apex frame) of real-world MiE videos. \textbf{(b)} Early frames in real-world macro-expression videos are used to obtain $\mathbf{z}^{\text{MaE}}$. The hyperparameters of choosing the frame indices are selected in Section~\ref{sec:exp:analysis}. \textbf{(c)} We specify an emotion type (\eg, sad) and then the AU distribution from the Expert Mapping table~\cite{du2014compound}, which determine the activated AU entries. 
        Then we assign activated AU entries with intensity values (red bars) and others with $0$. The hyperparameters of constraining intensity values are experimented in Section~\ref{sec:exp:analysis}.}
        \label{figure: compute_three_AUs}   
    \end{center}
\end{figure*}

\textbf{AUs defined by expert knowledge.}
Studies reveal strong relationships between AUs and emotions~\cite{du2014compound,ekman1997face,polikovsky2013facial}. Some explicitly summarize the posterior probability of each AU entry being activated for each emotion label: $P(z_i>0|y)$, where $z_i$ indicates the $i$-th entry of AU vector $\mathbf{z}$. The posterior probabilities, for simplicity, are usually modeled with a Bernoulli distribution~\cite{du2014compound}, \ie, $P(z_i>0|y) = p$ and $P(z_i=0|y) = 1 - p$. We find the AU distribution summarized by experts another effective source of AUs for synthesizing trainable MiEs. 

We use the expert knowledge mainly to find the apex AUs, 
where we resort to a mapping table \cite{du2014compound} that describes the aforementioned posterior probabilities. Given an emotion label, when generating the apex AUs $\mathbf{z}_{a}^{\text{exp}}$, we first decide which entries in $\mathbf{z}_{a}^{\text{exp}}$ should be activated ($>0$) by drawing samples from the Bernoulli distribution. We then determine the intensities of the activated entries by randomly sampling from a uniform distribution with a fixed interval $[\mu, \nu]$. The selection of hyperparameters $\mu, \nu$ is briefly discussed in Section~\ref{sec:exp:analysis}.
On the other hand, for the onset AUs $\mathbf{z^{\text{exp}}_o}$, we set them to zero vectors, which means that no action unit is activated, thus representing a neutral face.
Examples of how to compute the above three types of AUs are provided in Fig.~\ref{figure: compute_three_AUs}.

\emph{Discussion.} We use three basic expression categories (\texttt{positive}, \texttt{negative}, \texttt{surprise}) when synthesizing MiE-X, because these three classes form the largest common intersection between the label sets from the three sources. If we could have more fine-grained label space, it would be interesting to further explore how the label space affects the training quality of MiE-X.

\begin{figure*}[t] 
    \centering
    \begin{center}
        \includegraphics[width=\textwidth]{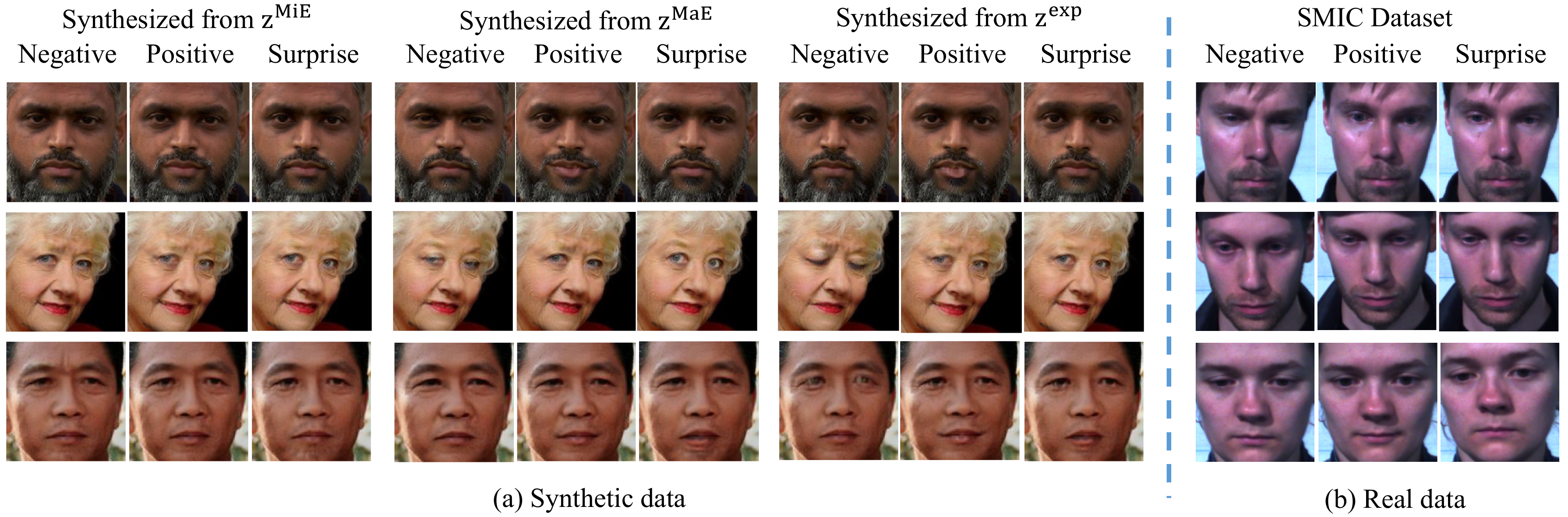}
        \caption{Examples of MiE apex frames from \textbf{(a)} synthetic (MiE-X) and \textbf{(b)} real-world (the SMIC dataset \cite{li2013spontaneous}) micro-expression data. In (a), we show three columns of synthesized MiE apex frames corresponding to three types of Action Units (AUs), \ie, $\mathbf{z}^{\text{MiE}}$,$\mathbf{z}^{\text{MaE}}$, $\mathbf{z}^{\text{exp}}$ described in Section \ref{sec:AUs}. Both real-world data and synthetic data the shown under classes labels \texttt{positive}, \texttt{negative}, and \texttt{surprise}.}
        \label{figure: Different_data}
    \end{center}
\end{figure*}

\subsection{The MiE-X dataset} \label{method:MIEX}
With the above three types of AUs and a large pool of in-the-wild faces, we eventually are able to synthesize a large-scale MiE recognition dataset, coined MiE-X. MiE-X contains 5,000 identities, each with 9 MiE samples\footnote{For each ID and each of the three classes \texttt{positive}, \texttt{negative}, and \texttt{surprise}, we generate three MiE samples corresponding to three types of AUs. Each sample has an onset and an apex frames, totaling 9 MiE samples and 18 frames per ID.}
, resulting in 45,000 samples in total. To our knowledge, MiE-X is the first large-scale MiE dataset and is more than two orders of magnitude larger than existing real-world MiE datasets. Visualization of the generated apex frames in MiE-X is provided in Fig.~\ref{figure: Different_data}; comparisons with existing MiE datasets are illustrated in Fig.~\ref{fig:compare_datasets}.

The strength of MiE-X as training data comes from its diversity 
in identity and MiE patterns\footnote{We also acknowledge  GANimation that provides us with realistic facial images.}. For instance, it contains 5,000 human identities, encouraging models to learn identity-invariant expression features. At the same time, the three sources of AUs are complementary, provide a wide range of AU values, and sometimes have random AU perturbations. MiE-X alleviates overfitting risks and allows algorithms to consistently improve their accuracy. 


\section{Experiment}\label{sec:exp}

\subsection{Experimental setups}\label{sec:exp:setup}

\textbf{Baseline classifiers.}\label{sec:exp:classifier}
Two image-based MiE recognition methods are mainly evaluated in this paper:
the \textbf{Branches}~\cite{Liu2018A} and \textbf{ApexME}~\cite{li2018can}. Both are trained for 80 epochs. More details are provided in supplementary materials. 

\textbf{Real-world datasets.} 
We report experimental results on commonly-used real-world datasets:
\textbf{CompMiE~\cite{see2019megc}}, \textbf{MMEW} and \textbf{SAMM}. CompMiE is proposed by the MiE recognition challenge MEGC2019~\cite{see2019megc} which merges three existing real MiE datasets into one. The three component datasets are CASME II~\cite{yan2014casme}, SAMM~\cite{davison2016samm,davison2018objective}, and SMIC~\cite{li2013spontaneous}, respectively. CompMiE has the same label space (Section~\ref{sec:AUs}) as MiE-X and consists of 442 samples from 68 subjects in total.
MMEW and SAMM have 234 and 72 samples, respectively, and their label spaces are different with MiE-X\footnote{Label space of MMEW: \texttt{happiness}, \texttt{surprise}, \texttt{anger}, \texttt{disgust}, \texttt{fear}, \texttt{sadness}; Label space of SAMM: \texttt{happiness}, \texttt{surprise}, \texttt{anger}, \texttt{disgust}, \texttt{fear}.}. 
The MaE dataset \textbf{CK+}~\cite{lucey2010extended} is a commonly used real-world MaE dataset containing 327 videos. Its label space is also merged into the same one as CompMiE. 
When generating MiE-X (see Section~\ref{sec:Synthesis}), we extract $\mathbf{z}^{\text{MiE}}$ and $\mathbf{z}^{\text{MaE}}$ from CompMiE and CK+, respectively.


\textbf{Evaluation protocols.} 
We use subject-wise $k$-fold cross-validation, commonly performed in the community \cite{ben2021video,li2018can,khor2018enriched}. Specifically, when real-world data are used in testing, we split them into $k$ subsets. Each time, we use $k-1$ subsets for training and the rest $1$ subset for testing. The average accuracy of the $k$ tests is reported. For CompMiE, $k=3$; for MMEW and SAMM, $k=5$. To evaluate the effectiveness of MiE-X, we replace real training sets (\ie, $k-1$ subsets) with MiE-X when MiE-X is used for direct deployment. Note that, for each fold, MiE-X samples whose AUs (\ie, $\mathbf{z}^{\text{MiE}}$) are computed from real MiE samples in the test subset will not be used in training.
If MiE-X is used for pre-training, where a fine-tuning stage is required, the $k-1$ subsets will be used for fine-tuning.
Other real-world datasets (\eg, MMEW, SMIC) are also used for pre-training to form comparisons with MiE-X\footnote{We discard those samples in real-world datasets that overlap with the test subset.} 
Experiment is categorized as follows.


\begin{itemize}
   \item {Pre-training with MiE-X (or other competing datasets) and fine-tuning on target training set.} We adopt this setting especially when the source domain has a different label space from the target domain. 
   \item {Training (or fine-tuning) with MiE-X (or other competing datasets) followed by direct model deployment.} If the target domain and training dataset share the same label space, models obtained from the training set can be directly used for inference on the target test set. 
\end{itemize}



\textbf{Metrics.} We mainly use unweighted F1-score (UF1) and unweighted average recall (UAR) \cite{see2019megc}. UF1 and UAR indicate the average F1-score and recall, respectively, over all classes. 
We also report the conventional recognition rate on the MMEW~\cite{ben2021video} and SAMM~\cite{davison2018objective} datasets to compare with the state of the art. By default, we run each experiment ($k$-fold cross-validation) 3 times and report the mean and standard variance of the results in the last epoch. Moreover, we provide the best accuracy among all epochs for reference (Table~\ref{table:SOTA}).

\subsection{Effectiveness of the Synthetic Database}\label{sec:exp:main_results}

\begingroup
\setlength{\tabcolsep}{3.9pt} 
\begin{table}[t]
\centering
\footnotesize
\caption{Effectiveness of MiE-X in model (pre-)training. 
Models are pre-trained using MiE-X or other real-world datasets and then fine-tuned on real-world training data \ie, CompMiE, or the combination of CompMiE and CK+~\cite{lucey2010extended}. UF1 (\%) and UAR (\%) are reported on the CompMiE dataset after three-fold cross-validation. ApexME~\cite{li2018can} and Branches~\cite{Liu2018A} are used as baselines. We observe consistent accuracy improvement when models are pre-trained with MiE-X. In addition, when directly deploying the MiE-X pretrained model, the accuracy is also competitive.}
\begin{tabular}{c|cc|cc|cc} \toprule
    \multicolumn{1}{c|}{{Pre-training}} & \multicolumn{2}{c|}{{Fine-tuning}} & \multicolumn{2}{c|}{ApexME~\cite{li2018can}}  & \multicolumn{2}{c}{Branches~\cite{Liu2018A}}\\ 
    \cline{2-7}
     MiE data &  CompMiE & CK+ & UF1 & UAR & UF1 &UAR \\
    \midrule   
     - & \cmark & & 41.8 $\pm$ 0.7 & 41.9 $\pm$ 0.7 & 43.6 $\pm$ 0.5  & 44.6 $\pm$ 0.6 \\ 
     - & \cmark & \cmark & 45.0 $\pm$ 0.5 & 45.5 $\pm$ 1.0 & 45.2 $\pm$ 0.5 & 47.0 $\pm$ 0.6 \\ 
     SMIC~\cite{li2013spontaneous} & \cmark &  & 45.0 $\pm$ 1.7  &  44.8 $\pm$ 1.9  & 42.8 $\pm$ 0.8 & 41.4 $\pm$ 0.9  \\ 
     CASME~\cite{yan2014casme} & \cmark  &  & 44.0 $\pm$ 1.2  &  45.1 $\pm$ 0.5  & 40.7 $\pm$ 0.9 &  41.4 $\pm$ 0.9  \\ 
     SAMM~\cite{davison2016samm} & \cmark  &  & 43.7 $\pm$ 0.7  & 42.8 $\pm$ 0.5  & 42.3 $\pm$ 1.4 &  42.9 $\pm$ 1.7  \\ 
     
     MMEW~\cite{ben2021video} & \cmark  &  & 43.3 $\pm$ 0.8  &  44.4 $\pm$ 1.2  & 43.3 $\pm$ 1.3 &  44.1 $\pm$ 1.5  \\ 
     \cline{1-7}
     MiE-X & & & 45.2 $\pm$ 0.5 & 46.3 $\pm$ 0.5 & 47.7 $\pm$ 0.5 & 48.9 $\pm$ 0.8 \\
     MiE-X & \cmark  &  & 46.9 $\pm$ 0.9 & \textbf{48.3} $\pm$ 0.9 & 50.7 $\pm$ 0.9 & 52.1 $\pm$ 1.4 \\ 
     MiE-X & \cmark & \cmark & \textbf{47.0} $\pm$ 0.8 & 48.2 $\pm$ 0.4 & \textbf{52.3} $\pm$ 0.7 & \textbf{52.3} $\pm$ 0.4 \\
     \bottomrule
\end{tabular}
\label{table:Fine_Tuning}
\end{table}
\endgroup

\textbf{Effectiveness of MiE-X in training models for direct deployment.} MiE-X has the same label space with CompMiE. 
So models trained with MiE-X can be directly evaluated on the CompMiE. In Table \ref{table:Fine_Tuning}, ApexME and Branches trained with MiE-X alone produce an UF1 of 45.2\% and 47.7\%, respectively, which outperforms the training set composed of CompMiE and CK+. 



\textbf{Effectiveness of MiE-X in model pre-training.} First, when using MiE-X for model pre-training, we observe consistent improvement over not using it (Table~\ref{table:Fine_Tuning}). For example, when we perform fine-tuning on CompMiE using the ApexME method, pre-training with MiE-X brings 5.1\% and 7.1\% improvement in UF1 and UAR, respectively, over not using MIE-X. 
Second, we compare MiE-X with existing datasets (\ie, SMIC, CASME, SAMM, and MMEW) of their effectiveness as a pre-training set, on which we train the baseline MiE classifiers (\ie, ApexME, Branches). We do three-fold cross-validation on CompMiE. For each fold, we use the dataset (e.g., SMIC) we would like to evaluate as the pre-training data. Samples are removed from the training set if they also appear in the test subset of CompMiE in the current fold. Then we fine-tune the model on the training subset of CompMiE.  Results are shown in both Table~\ref{table:Fine_Tuning} and Fig.~\ref{fig:compare_datasets}. We observe that the model pre-trained on MiE-X significantly outperforms those pre-trained on other datasets. For instance,  when we pre-train Branches on MiE-X, the final fine-tuning results on CompMiE in UF1 and UAR are 7.4\% and 8.0\% higher than using MMEW as the pre-training data. This phenomenon validates the effectiveness of our dataset and the proposed synthesis procedure.

\begingroup
\setlength{\tabcolsep}{6.4pt} 
\begin{table}[t]
\centering
\footnotesize
\caption{Comparison with the state-of-the-art MiE recognition methods on MMEW and SAMM datasets. We re-implement ApexME, Branches and DTSCNN, which are pretrained with either ImageNet or MiE-X (grey). We report the mean recognition accuracy (\%) and standard variance. 
$\dagger$ donates vide-based methods. ``Last'' means test result in the last epoch, and ``Best'' refers to the best accuracy among all epochs.}
{
\begin{tabular}{l|cccc} \toprule
    \multirow{2}{*}{Methods} & \multicolumn{2}{c}{MMEW} & \multicolumn{2}{c}{SAMM}\\
    \cline{2-5} 
    & Last & Best  & Last & Best \\
    \midrule 
    FDM~\cite{xu2017microexpression} &  \quad34.6 & $-$ & 34.1 & $-$\\
    LBP-TOP~\cite{zhao2007dynamic} & \quad 38.9 & $-$ & 37.0 & $-$\\
    DCP-TOP~\cite{ben2018learning} &  \quad42.5 & $-$ & 36.8 & $-$\\
    ApexME~\cite{li2018can} &  \quad 48.5 $\pm$ 0.6 & 58.3 $\pm$ 0.9 & 41.3 $\pm$ 0.6 & 54.9 $\pm$ 0.7 \\
    \rowcolor{mygray}
    ApexME \textbf{+ MiE-X} & \quad 55.9 $\pm$ 2.0 &  61.4 $\pm$ 0.8  & 46.4 $\pm$ 0.7 & 60.3 $\pm$ 1.1 \\
    Branches~\cite{Liu2018A}  & \quad  50.1 $\pm$ 0.6 & 58.3 $\pm$ 0.6 & 44.5 $\pm$ 0.7 & 53.3 $\pm$ 0.5 \\
    \rowcolor{mygray}
    Branches \textbf{+ MiE-X} &  \quad 56.8 $\pm$ 1.1 & 61.5 $\pm$ 1.0 &  48.7 $\pm$ 1.0 & 56.3 $\pm$ 0.8 \\
    \midrule 
    TLCNN$^\dagger$~\cite{wang2018micro} & $-$ &\quad 69.4 & $-$ & 73.5 \\
    DTSCNN$^\dagger$~\cite{peng2017dual} & 60.9 $\pm$ 1.3 & \quad 71.1 $\pm$ 1.1 & 51.6 $\pm$ 1.8 & 60.6 $\pm$ 1.1 \\
    \rowcolor{mygray}
    DTSCNN$^\dagger$ \textbf{+ MiE-X} & 63.1 $\pm$ 1.0 & \quad 74.3 $\pm$ 0.5 & 55.5 $\pm$ 1.4 & 73.9 $\pm$ 0.9 \\
    \bottomrule
\end{tabular}
}
\centering
\label{table:SOTA}
\end{table}
\endgroup

\begin{figure*}[t] 
    \centering
    \begin{center}
        \includegraphics[width=\textwidth]{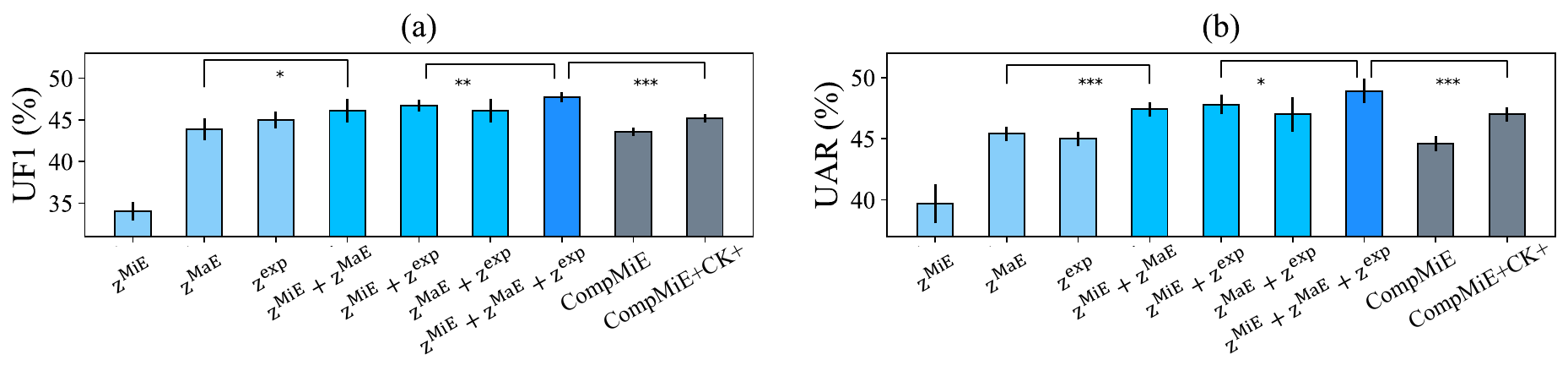}
        \caption{Comparing training effectiveness of real-world data and various synthetic datasets sourced from different combinations of AUs. We compare UF1 \textbf{(a)} and UAR \textbf{(b)} on CompMiE. ``n.s.'' means the difference is {not statistically significant} ($i.e., p$-value $>$ 0.05).  $*$ denotes {statistically significant} ($i.e., 0.01 < p$-value $< 0.05$). $**$ and $***$ mean {statistically very significant} ($i.e., 0.001 < p$-value $< 0.01$) and {statistically extremely significant} ($i.e., p$-value $< 0.001$), respectively. We observe decreased accuracy if we remove any of the three types of AUs. When all the three types are used for database creation, both UF1 and UAR exceed results obtained by training on real-world data, with very high statistical confidence.}
        \label{figure: ingredients} 
    \end{center}
\end{figure*}

\textbf{Positioning within the state of the art.} 
We follow a recent survey~\cite{ben2021video} and compare with the state of the art on two datasets, MMEW~\cite{ben2021video} and SAMM~\cite{davison2018objective}, all under 5-fold cross validation. 
Results are summarized in Table~\ref{table:SOTA}. 
We re-implemented three baselines (ApexME, Branches and DTSCNN),  pretrained  on either ImageNet or MiE-X. 
To pretrain the video-base method DTSCNN, we use a simple variant of MiE-X where each sample has multiple frames. Specifically, when computing $\mathbf{z}^{\text{MiE}}$ and $\mathbf{z}^{\text{MaE}}$, we extract AUs for all the frames between the onset and apex frames. All these extracted AUs are used for frame generation. For $\mathbf{z}^{\text{exp}}$, we linearly interpolate 8 AU vectors between the onset and apex AU vectors, thus generating 10 frames per sample.

Table 2 clearly informs us that MiE-X pre-training improves the accuracy of all the three methods. 
Importantly, when MiE-X is used for pre-training, MiE recognition accuracy is very competitive: DTSCNN achieves accuracy (best epoch) of 74.3 $\pm$ 0.5 \% and 73.9 $\pm$ 0.9 \% on MMEW and SAMM, respectively.





\subsection{Further Analysis} \label{sec:exp:analysis}
All experiments in this section are performed on the Branches baseline \cite{Liu2018A}. 

\textbf{Comparisons of various AU combinations.} 
Fig.~\ref{figure: ingredients} evaluates various AU combinations on CompMiE. We have the following observations. 
\textbf{First}, none of the three types of AUs are dispensable. We observe that the best recognition accuracy is obtained when all three types of AUs are used, which outperforms training with CompMiE+CK+ by 1.7\% and 2.0\% in UF1 and UAR, respectively. Importantly, if we remove any single type of AUs, the UF1 and UAR scores decrease. For example, when removing $\mathbf{z}^{\text{MiE}}$, $\mathbf{z}^{\text{MaE}}$, $\mathbf{z}^{\text{exp}}$ one at a time, the decrease in UF1 score is 1.6\%, 1.0\% and 1.6\%, respectively. 

\textbf{Second}, using two types of AUs outperforms using only a single type with statistical significance. For example, when using $\mathbf{z}^{\text{MiE}}$ and $\mathbf{z}^{\text{MaE}}$, UF1 is higher than using $\mathbf{z}^{\text{MaE}}$ alone by 2.15\%. In fact, the three AU types come from distinct and trustful sources, allowing them to be complementary and effective. This also explains why all three AU types are better than any combination of two. 

\textbf{Third}, when using a single type of AUs, we find $\mathbf{z}^{\text{MaE}}$ or $\mathbf{z}^{\text{exp}}$ produces much higher UF1 and UAR than $\mathbf{z}^{\text{MiE}}$. Their superiority could be explained by their diversity.  
Compared with $\mathbf{z}^{\text{MiE}}$, MiEs generated from $\mathbf{z}^{\text{MaE}}$ and $\mathbf{z}^{\text{exp}}$ are much more diverse. Specifically, when constructing $\mathbf{z}^{\text{MaE}}$, the index of apex frame is randomly drawn from a range $\lfloor\alpha \times n\rfloor$ and $\lfloor\beta \times n\rfloor$. Similarly, the randomness of AU intensities is also introduce by hyperparameter $\mu$ and $\nu$ when generating $\mathbf{z}^{\text{exp}}$. In contrast, the index of the apex frame is fixed when constructing $\mathbf{z}^{\text{MiE}}$. 

\textbf{Lastly}, we compare results that employ two real-world training datasets. The first is CompMiE, described as in Section~\ref{sec:exp:setup}, and the second is a combination of CompMiE and CK+. It is shown that CompMiE + CK+ outperforms CompMiE by an obvious margin, suggesting that \emph{early-stage of MaEs highly correlate with MiEs}. 
These results motivated us to mine effective AUs ($\mathbf{z}^{\text{MaE}}$) from MaEs.

\begin{figure}[t]
    \centering
    \begin{center}
        \includegraphics[width=1\textwidth]{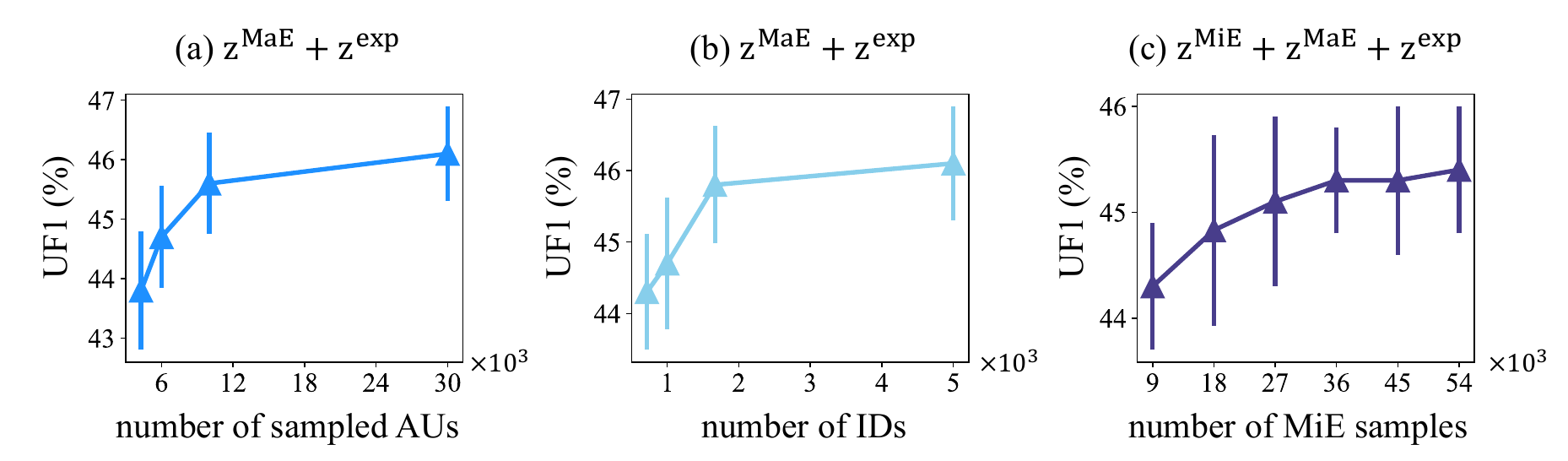}
        \caption{\textbf{(a)-(b)}: Impact of the number of AU triplets \textbf{(a)}, IDs \textbf{(b)} and MiE samples \textbf{(c)}. In \textbf{(a)-(b)}, we use $\mathbf{z}^{\text{MaE}}$ and $\mathbf{z}^{\text{exp}}$ for database synthesis, while in \textbf{(c)} all three types AUs are used. We employ the Branches method \cite{Liu2018A}. When we gradually increase the numbers, the three-fold cross validation accuracy (UF1, \%) on CompMiE first improves and then remains stable in all the three subfigures. }
        \label{fig:numbers_ID_AU_img}
    \end{center}
\end{figure}

\textbf{Impact of the number of AUs, IDs and MiE samples in MiE-X.} For MiE-X, the IDs, AUs and MiE samples are all important, and 
we now investigate how their quantities influence MiE recognition accuracy by creating MiE-X variants with different numbers of IDs, AU triplets and samples. 
Here, please note that the diversity is highly relevant to the number of distinct IDs/AUs/samples, so sometimes we use number and diversity interchangeably. 
When studying AU and ID diversity, we set the AU combination to be $\mathbf{z}^{\text{MaE}} + \mathbf{z}^{\text{exp}}$ because their diversity can be easily changed by specifying the number of sampling times from the uniform distributions (refer Section~\ref{sec:AUs}). When investigating the number of MiE samples, we use all three types of AUs. 

To evaluate the influence of \textbf{AU} diversity, we set the number of MiE samples and IDs to 30,000 and 5,000 {(6 samples per ID)}, respectively in all the dataset variations. 
The AU diversity can be customized by allowing multiple identities to share the same AU triple. Specifically, the number of AU triplets is set to 4,000, 6,000, 10,000, 30,000 and 
From the experimental results in Fig.~\ref{fig:numbers_ID_AU_img} (a), we observe the effectiveness of synthetic data generally increases when AU diversity is improved. For example, the UF1 score increases by 1.8\%, when the number of distinct AU triplets increases from 4,000 to 10,000. When the number of AUs is greater than 10,000, the curve reaches saturation.

To study the diversity of \textbf{IDs}, we fix the number of MiE samples and AU triplets in MiE-X to be 30,000. We set the ID number as 700, 1,000, 1,700, and 5,000, 
achieved by randomly selecting face images from the EmotionNet~\cite{fabian2016emotionet} dataset\footnote{Note that each image in EmotionNet usually denotes a different identity.}.  
In this experiment, an ID {generates more than 6} MiE samples using AU triplets randomly drawn from the pool of 30,000. Results in Fig.~\ref{fig:numbers_ID_AU_img} (b) show that more IDs leads to a higher recognition accuracy. For example, UF1 of synthetic dataset increases from 44.4\% to 45.8\% when the number of IDs increases from 700 to 1,700. When the number of IDs exceeds 1,700, the curve becomes stable. 


To study the impact of the number of \textbf{MiE samples},  
we fix the number of AU triplets to 9,000 and the number of IDs to 1,000. We then gradually increase the generated samples from 9,000 to 54,000 by reusing more AU triplets on each ID. 
Experimental results are shown in Fig.~\ref{fig:numbers_ID_AU_img} (c). We find the effectiveness of the synthetic training set generally increases when more samples are included and that curve becomes flat when the number of samples are greater than 36k. 
For example, the UF1 is improved by 1.0\%, when the number of samples increases from 9k to 36k. When the number of samples increases from 36k to 54k, there is a slight UF1 improvement of 0.2\%. 
This observation is expected because when the number of IDs and AUs are fixed, the total information contained in the dataset is constrained. From the above experiments, we conclude that MiE-X benefits from more AUs, IDs and samples within a certain range.

\begin{wraptable}{r}{0.5\textwidth}
			\centering
			\setlength{\tabcolsep}{1mm}{
			\small
 			\caption{
			Performance comparison between training with and without side faces. Evaluation is on the CompMiE dataset.
			}
			\label{tab:pose}
			\resizebox{0.4\textwidth}{!}{
				\begin{tabular}{|c|c|c|}
					\hline	
				     &  \emph{w/} side & \emph{w/o} side \\
					\hline 
					UF1 (\%)  & 47.7 $\pm$ 0.5 & 47.4 $\pm$ 0.8 \\

					\hline 
				\end{tabular}
			}}
\end{wraptable}

\textbf{Impact of face poses.} 
We use 5,000 IDs with frontal faces to synthesize a training set variant which is compared with MiE-X composed of faces of various poses. To find the frontal faces, we manually select 10 frontal faces in the  EmotionNet dataset as queries and for each search for 500 faces with similar facial landmarks detected by a pretrained MTCNN landmark detector \cite{7553523}.
{Table~\ref{tab:pose} summarizes the results on CompMiE, where we do not observe obvious difference between the two training sets.} 
This can possibly be explained by the fact that real-world MiE datasets mostly contain frontal faces 
collected in laboratory environments. Therefore, pose variance in MiE-X may not significantly influence performance on existingtests. Nevertheless, we speculate using various poses to generate MiE-X would benefit MiE recognition in uncontrolled environments.

\begin{wrapfigure}{r}{0.5\textwidth}
    \centering
        \includegraphics[width=0.5\textwidth]{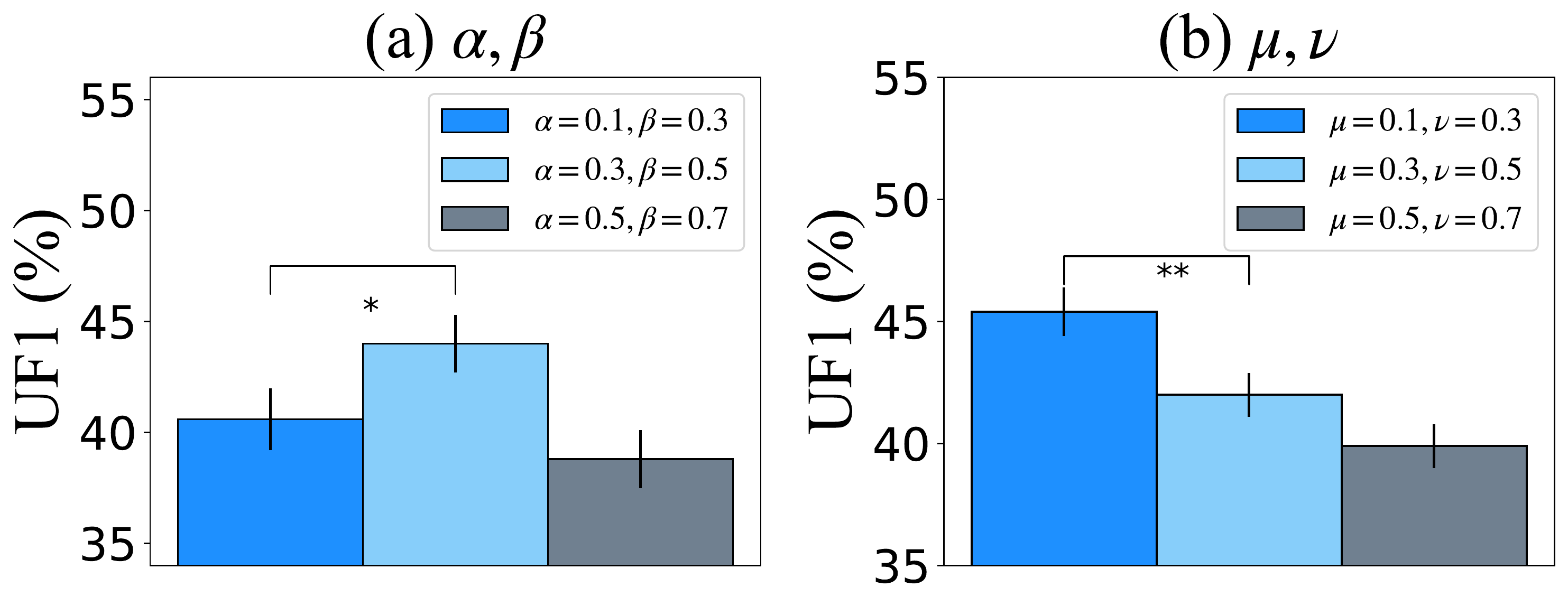}
        
        \caption{Impact of hyperparameters in computing $\mathbf{z}^{\text{MaE}}$ and $\mathbf{z}^{\text{exp}}$. UF1 (\%) on the CompMiE dataset is reported in each sub-figure. \textbf{(a):} MiE-X is composed by $\mathbf{z}^{\text{MaE}}$ only. Three groups of $\alpha$ and $\beta$ values are tested. \textbf{(b):} MiE-X is made from $\mathbf{z}^{\text{exp}}$ only. Three groups of $\mu$ and $\nu$ are investigated. $*$ and  $**$ have the same meaning as Fig.~\ref{figure: ingredients}. }
        \label{fig:exp:hyperparameter} 

\end{wrapfigure}

\textbf{Analysis of other hyperparameters.}
Due to the lack of validation data in real-world MiE datasets, we mostly used prior knowledge and intuition to choose the hyperparameters. Specifically, we chose $\alpha = 0.3$, $\beta = 0.5$ and $\mu = 0.1$, $\nu = 0.3$ in experiments.  Here, we briefly analyze these two sets of hyperparameters involved in the AU computation on CompMiE using cross-validation. $[\alpha,\beta]$ is the interval from which the apex frames for computing $\mathbf{z}^{\text{MaE}}$ are randomly selected. 
Specifically, we analyze three options: ($\alpha = 0.1$, $\beta = 0.3$), ($\alpha = 0.3$, $\beta = 0.5$) and ($\alpha = 0.5$, $\beta = 0.7$). The number of identities is 5,000. Recognition accuracy of the three options is given by Fig.~\ref{fig:exp:hyperparameter} (a), where $\alpha = 0.3$, $\beta = 0.5$ produces the highest UF1 score. This result is in accordance with our intuition: the first $30\%$ to $50\%$ frames of an MaE would be more similar to an MiE. 

$[\mu$, $\nu]$ is the interval from which the intensities of expert-defined AUs are uniformly sampled. Similarly, we analyze three options, \ie, ($\mu = 0.1$, $\nu = 0.3$), ($\mu = 0.3$, $\nu = 0.5$) and ($\mu = 0.5$, $\nu = 0.7$).
This is inspired by observing AU coefficients of real MiEs: the intensity of each action unit is not large, \ie, $< 0.7$ in most cases, because micro-expressions have {subtle} facial muscle movements. Results are shown in Fig.~\ref{fig:exp:hyperparameter} (b): the intensity range $[0.1, 0.3]$ is superior. 
Because the highest value of an MaE AU is $1.0$, the value of $[\mu$,$\nu]$ delivers another intuitive message: {facial AU intensities of MiEs are around $10\%$ to $30\%$ those of MaEs.}

\subsection{Understanding of MiEs: A Discussion}

\textbf{MiEs generalize across faces.} AUs extracted from real MiEs provide closest resemblance to true MiEs and are thus indispensable. These AUs $\mathbf{z}^\text{MiE}$ are generalizable because they can be transplanted to faces of different identities. The fact that a higher number of face identities generally leads to a higher accuracy indicates the benefit of adding AUs $\mathbf{z}^\text{MiE}$ to sufficiently many faces to improve MiE recognition towards identity invariance. 

\textbf{Early-stage MaEs resemble real MiEs.} To our knowledge, we make very early attempt to leverage MaEs for MiE generation. Although the two types of facial expressions differ significantly in their magnitude of facial movement, we find AUs in initial stages of MaEs are effective approximations to those in MiEs. 

\textbf{Expert knowledge is transferable to MiEs.} 
While AUs annotated by experts are used to describe MaEs, we find expert AUs with reduced magnitude are effective in synthesizing MiEs. We therefore infer from a computer vision viewpoint that MiEs are related to normal expressions but with lower intensity. Moreover, by examining the complementary nature of the three types of AUs, we infer that expert knowledge adds some useful computational cues, which do not appear in MaEs and real MiEs but can be humanly defined. Nevertheless, our work is limited in that the psychological aspects of MiEs are not considered, which will be studied in future with cross-disciplinary collaborations.


\section{Conclusion}
This paper addresses the data lacking problem in MiE recognition. 
An important contribution is the introduction of a large-scale synthetic dataset, MiE-X, with standard emotion labels to improve MiE model training. 
In the synthesis protocol, we feed faces in the wild, desired emotion labels and AU triplets (our focus) to a generation model. 
Specifically, sourced from real MiEs, early-stage MaEs, and expert knowledge, three types of AUs are identified as useful and complementary to endorse an effective protocol. This understanding of the role of AUs in effective MiE synthesis is another contribution of this work. 
Experiment on real-world MiE datasets indicates MiE-X is a very useful training set: models (pre-)trained with MiE-X consistently outperform those (pre-)trained on real-world MiE data.
In addition, this paper reveals some interesting computational properties of MiEs, which would be of value for further investigation.

\clearpage
%
%

\bibliographystyle{splncs04}

\end{document}


\pagestyle{headings}
\mainmatter
\def\ECCVSubNumber{2280}  

\title{Supplementary Material for \\ ``How to Synthesize a Large-Scale and Trainable Micro-Expression Dataset?"}



\titlerunning{How to Synthesize a Large-Scale and Trainable Micro-Expression Dataset?}
%
\author{Yuchi Liu \inst{1} \and
Zhongdao Wang\inst{2} \and
Tom Gedeon\inst{1} \and
Liang Zheng \inst{1}}
\authorrunning{Y. Liu, et al.}
%
\institute{Australian National University, Canberra, Australia, \\
\email{\{firstname.lastname\}@anu.edu.au} \and
Tsinghua University, Beijing, China
\email{wcd17@mails.tsinghua.edu.cn}\\
\url{https://github.com/liuyvchi/MiE-X}
}
\maketitle

In the supplementary material, we 1) visualize the values of computed action units of $\mathbf{z}^{\text{MiE}}$ and $\mathbf{z}^{\text{MaE}}$, 2) show the effectiveness of MiE-X variants using different dataset as ID sources and different datasets as AU sources and 3) the implementation details of models involved in this paper.

\section{More Analysis on $\mathbf{z}^{\text{MiE}}$ and $\mathbf{z}^{\text{MaE}}$. }

Fig.~\ref{fig:bar_au_comparision} visualizes the averaged values of Action Unit (AU) vectors of $\mathbf{z}^{\text{MiE}}$ (extracted from \textit{micro-expressions}) and $\mathbf{z}^{\text{MaE}}$ (extracted from \textit{early-stage macro-expressions}), respectively, under different micro-expression categories. 
The setting of AU numbers is same to that used in GANimation~\cite{pumarola2018ganimation}~\footnote{17 representative AU numbers are utilized (\textit{i.e.,} AU1, AU2, AU4, AU5, AU6, AU7, AU9, AU10, AU12, AU14, AU15, AU17, AU20, AU23, AU25, AU26, and AU45).}. 

\begin{center}
    \centering
    \includegraphics[width=\textwidth]{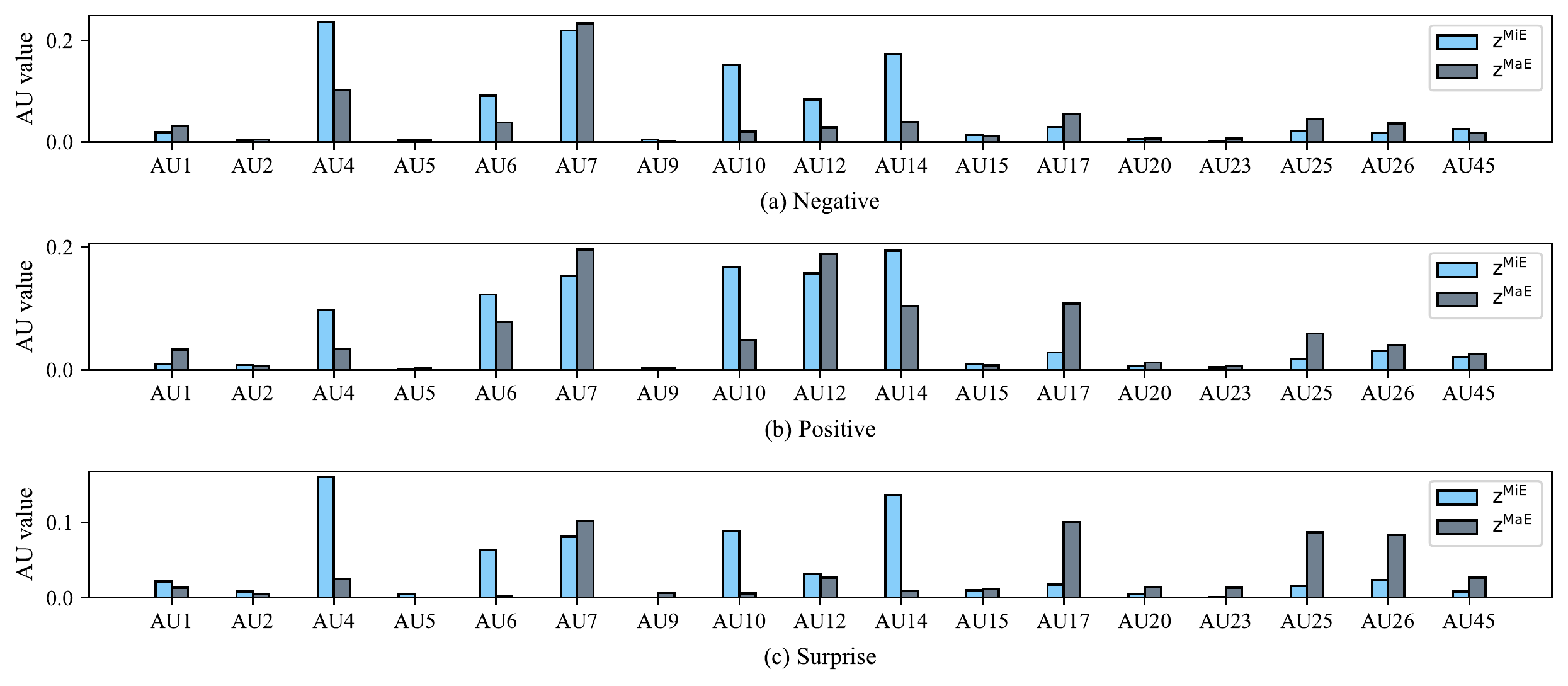}
    \captionof{figure}{Averaged values of 17 Action Units calculate on $\mathbf{z}^{\text{MiE}}$ (extracted from micro-expressions) and $\mathbf{z}^{\text{MaE}}$ (extracted from early-stage macro-expressions), respectively, under different emotion categories (\ie, \texttt{negative}, \texttt{positive} and \texttt{surprise}). For each category, we have two observations:  1) $\mathbf{z}^{\text{MiE}}$ and $\mathbf{z}^{\text{MaE}}$ share similar trends among different AU. For example, both $\mathbf{z}^{\text{MiE}}$ and $\mathbf{z}^{\text{MaE}}$ have high values of AU7 for Negative. It means they are depicting for the same AU of each kind of expressions, 2) $\mathbf{z}^{\text{MiE}}$ and $\mathbf{z}^{\text{MaE}}$ have different values, which means they are complementary to each other.}
    \label{fig:bar_au_comparision}
\end{center}


There are two major findings. First, $\mathbf{z}^{\text{MiE}}$ and $\mathbf{z}^{\text{MaE}}$ show similar 
activation patterns when they have same emotion labels. For example, when the emotion category is positive, both types of AUs have relatively large values in AU7, AU12 and AU14. Second, 

$\mathbf{z}^{\text{MiE}}$ and $\mathbf{z}^{\text{MaE}}$ show very different average values for some AU numbers. For instance, their values in AU10, AU14, and AU17 are dissimilar for all the three categories. This suggests $\mathbf{z}^{\text{MiE}}$ and $\mathbf{z}^{\text{MaE}}$ complement each other to some extent. This might explain why using synthetic data generated from both $\mathbf{z}^{\text{MiE}}$ and $\mathbf{z}^{\text{Mae}}$ for MiE recognition learning achieves higher accuracy than only using one of them (refer Fig. 5 of the main paper).

\begin{wraptable}{r}{0.48\textwidth}
			\centering
			\setlength{\tabcolsep}{1mm}{
			\small
 			\caption{
			 Comparing two MiE-X variants with different ID sources. We report the three-foldd cross-validation results on the CompMiE dataset. EmotionNet and CelebA are compared. The Branches method \cite{Liu2018A} is used.
			}
				\begin{tabular}{|c|c|c|}
					\hline	
				    ID Source &  EmotionNet & CelebA \\
					\hline 
					UF1 (\%)  & 47.7 $\pm$ 0.5 & 47.2 $\pm$ 0.7 \\

					\hline 
				\end{tabular}
			}
			\label{tab:change_ID_source}
\end{wraptable}

\section{Synthesizing MiE-X with A Different ID Source}
In the main paper, we use the EmotionNet~\cite{fabian2016emotionet} dataset to sample face IDs. To analyze the generalization ability of the proposed MiE generation protocol on different face datasets, we use CelebA~\cite{liu2015faceattributes} to replace EmotionNet for ID sampling while controlling the numbers of AU triplets and IDs to be unchanged. The generated MiE-X variant whose IDs come from CelebA is evaluated on the CompMiE dataset by three-fold cross validation. Comparative results are shown in Table \ref{tab:change_ID_source}. From the results, we do not observe significant difference in UF1 score after we change the ID source. The results also suggest one of the major findings: MiEs generalize cross faces.

\section{Synthesizing MiE-X Using Different Datasets as AU Sources}
\begin{table} 
			\centering
			\setlength{\tabcolsep}{2mm}{
			\small
 			\caption{
			Performance comparison between MiE-X variants with different ID sources. We report the three-fold cross-validation results (UF1, \%) on the CompMiE dataset. The Branches method \cite{Liu2018A} is used.
			}
			\label{tab:change_AU_source}
			\resizebox{0.98\textwidth}{!}{
				\begin{tabular}{|c|cc|cc|cc|}
					\hline	
				     AU &  \multicolumn{2}{c|}{$\mathbf{z}^{\text{MiE}}$ source} & \multicolumn{2}{c|}{$\mathbf{z}^{\text{MaE}}$ source} & \multicolumn{2}{c|}{$\mathbf{z}^{\text{MiE}}$ + $\mathbf{z}^{\text{MaE}}$ source} \\
					\cline{2-7} 
					Source & CompMiE & MMEW & CK+ & Oulu & CompMiE + CK+ & MMEW + Oulu \\
					\hline	
                    UF1 (\%) & 34.0 $\pm$ 1.1 & 37.2 $\pm$ 1.4 & 43.9 $\pm$ 1.3 & 42.33 $\pm$ 1.4 & 46.1 $\pm$ 1.5 & 47.5 $\pm$ 0.7\\
					\hline 
				\end{tabular}
			}}
\end{table}

In the main paper, $\mathbf{z}^{\text{MiE}}$ and $\mathbf{z}^{\text{MaE}}$ are extracted from the CompMiE \cite{see2019megc} dataset and the CK+ \cite{lucey2010extended} dataset, respectively. It is intriguing to investigate the effectiveness of MiE-X when we change its AU sources. 
Specifically, to compute $\mathbf{z}^{\text{MiE}}$ and $\mathbf{z}^{\text{MaE}}$, we use MMEW~\cite{ben2021video} and Oulu~\cite{zhao2011facial}, respectively.  
The AU computation and data generation protocols remain the same. 
We compare MiE-X variants with different AU sources in Table~\ref{tab:change_AU_source}. We observe that our MiE synthesis protocol can still generate competitive MiE datasets with alternative AU sources. For example, the MiE-X variant with AUs from MMEW and Oulu achieves an UF1 score of 47.5\% on CompMiE, while UF1 of the same model trained on CompMiE is 43.6\% (refer Table 1 in the main paper).

\section{Implementation Details of GANimation}
This paper employs the GANimation method~\cite{pumarola2018ganimation} to synthesize MiEs. It is an image-to-image translation model for face expression manipulation. 
Given a face image $\mathbf{x}^\mathbf{z}$ with Action Units (AUs) $\mathbf{z}$ and a target AU set $\mathbf{z'}$, GANimation aims to learn a single mapping function $G : (\mathbf{x}^\mathbf{z}, \mathbf{z'}) \rightarrow \mathbf{x}^\mathbf{z'}$ such that the generated face image not only has the same identity as the original image but also manifests the target AUs. We follow the default settings of GANimation during training.  Specifically, we use Adam with a learning rate of 0.0001 and batch size 25.  We train for 30 epochs and linearly decay the rate to 0 over the last 10 epochs. The weight coefficients for loss terms are the same as those in the original paper. With a single GTX 2080TI GPU, two days are needed to train this model.

\section{Implementation Details of The Baseline Classifiers}
We adopt the Branches~\cite{Liu2018A} and the Apex \cite{peng2018macro} as the baseline methods for MiE recognition in the main paper. Branches won the first place in~\cite{see2019megc} has two branches which do not share weights. An MiE sample has an onset frame and an apex frame. The two frames are fed into the two branches (using ResNet-18~\cite{he2016deep} as backbones), respectively, and their embeddings after global average pooling are concatenated.
The classier has two fully connected (FC) layers with dimension 128 and 32, respectively. We use Adam with a learning rate of 0.0001 and batch
size 32. We train for 80 epochs on real-world data 30 epochs on MiE-X. Compared with Branches, Apex only has one CNN branch and takes the apex frame as the input. It also uses ResNet-18 as the backbone. The training strategy of Apex is the same as those in training Branches. With a single GTX 2080TI GPU, around 10 hours are needed to train the baseline classifiers.

%
%

\bibliographystyle{splncs04}